# Leveraging Large Language Models for Predictive Analysis of Human Misery

Bishanka Seal [1], Rahul Seetharaman [2], Aman Bansal [2], and Abhilash Nandy [1]

[1] Indian Institute of Technology Kharagpur: Kharagpur, West Bengal, IN
[2] UMass Amherst, Amherst, USA
bishankaseal@kgpian.iitkgp.ac.in

**Abstract.** This study investigates the use of Large Language Models (LLMs) for predicting human-perceived misery scores from natural language descriptions of real-world scenarios. The task is framed as a regression problem, where the model assigns a scalar value from 0 to 100 to each input statement. We evaluate multiple prompting strategies, including zero-shot, fixed-context few-shot, and retrieval-based prompting using BERT sentence embeddings. Few-shot approaches consistently outperform zero-shot baselines, underscoring the value of contextual examples in affective prediction. To move beyond static evaluation, we introduce the "Misery Game Show", a novel gamified framework inspired by a television format. It tests LLMs through structured rounds involving ordinal comparison, binary classification, scalar estimation, and feedback-driven reasoning. This setup enables us to assess not only predictive accuracy but also the model's ability to adapt based on corrective feedback. The gamified evaluation highlights the broader potential of LLMs in dynamic emotional reasoning tasks beyond standard regression. Code and data link: https://github.com/abhi1nandy2/Misery_Data_Exps_GitHub

## 1  Introduction

The quantification of emotional distress from natural language remains a challenging yet increasingly relevant problem in computational social science and affective computing. Traditional assessment of psychological well-being relies heavily on structured interviews, clinical diagnostics, and standardized surveys. While effective in controlled environments, these approaches are inherently resource-intensive, susceptible to subjective bias, and limited in scalability. With the proliferation of digital text as a medium for self-expression, there is significant potential to infer latent emotional states directly from unstructured language.

Predicting misery scores from textual descriptions has practical utility in several domains. In mental health diagnostics, automatic misery quantification can support scalable early-warning systems by flagging emotionally distressing language in social media or online therapy sessions. In customer service and crisis management, such predictions can prioritize responses based on the severity of distress. Furthermore, applications in interactive storytelling, AI safety, and empathetic chatbot design benefit from LLMs capable of assigning emotional weight to narrative elements. These examples highlight the importance of developing fine-grained, human-aligned emotional reasoning systems.

Recent advancements in Natural Language Processing (NLP), particularly the emergence of Large Language Models (LLMs) such as GPT-3.5, GPT-4, and GPT-4o, offer powerful tools for semantic understanding and contextual reasoning. These models, trained on vast and diverse corpora, have demonstrated state-of-the-art performance across a wide array of tasks, including sentiment analysis, commonsense inference, and zero-shot classification [2,8,10]. For example, GPT-4 has been shown to exhibit early signs of general intelligence through its consistent performance across reasoning, code generation, and language understanding benchmarks [2]. Building upon this, GPT-3.5 and GPT-3 models have revealed significant capability in both instruction following and zero-shot generalization [10]. The recent GPT-4o further expands on these findings by demonstrating enhanced efficiency and performance in instruction tuning and real-time interaction [8].

In addition to task-oriented reasoning, researchers have begun investigating the affective capabilities of LLMs. Recent studies show that large models can internalize fine-grained emotional signals and respond with affectively coherent outputs, despite not being explicitly trained for emotion modeling [3]. Moreover, such capabilities appear to emerge naturally with scale and pretraining diversity, as highlighted in contemporary affective cognition evaluations of LLMs [4].

The integration of LLMs into mental health research is another notable trajectory. Several works have shown promise in using transformer-based models to detect emotional distress, depression, and other affective states from user-generated text [5], [15]. These models, when applied with care to ensure ethical considerations, offer potential for scalable and passive screening in online platforms.

In this work, we investigate the viability of LLMs for the task of misery score prediction, wherein a scalar value (ranging from 0 to 100) is assigned to a natural language description of a real-world scenario. Unlike binary or ordinal sentiment classification tasks, this regression-based formulation captures fine-grained variations in human-perceived emotional intensity. The task necessitates a blend of commonsense reasoning, emotional cognition, and sensitivity to contextual cues.

3We consider three prompting paradigms to evaluate model performance:
(i) Zero-shot prompting, wherein the model infers misery scores without exposure to labeled examples,
(ii) Few-shot prompting, where representative (statement, score) pairs are embedded in the prompt, and
(iii) Retrieval-augmented prompting, which dynamically selects semantically similar examples using BERT-based sentence embeddings [11].

To further probe model adaptability and decision-making under feedback, we introduce a novel *Misery Game Show Simulation*. Inspired by the television format *The Misery Index* [7], this gamified framework presents sequential prediction tasks across multiple rounds, with optional feedback after each iteration. The simulation is designed to assess both static regression accuracy and dynamic learning behavior under iterative supervision.

The contributions of this study are twofold. First, we benchmark LLM performance on a continuous misery prediction task under diverse prompting regimes. Second, we explore the capacity of LLMs to refine their predictions through feedback-driven reasoning in a simulated interactive environment. Empirical results indicate that few-shot prompting, particularly with semantically coherent context, substantially improves prediction accuracy over zero-shot baselines. Moreover, the feedback-augmented setting reveals measurable gains in adaptive learning, suggesting that LLMs possess a degree of flexibility in modeling subjective human evaluations.

## 2 Dataset Description

The dataset used in this study comprises 516 textual descriptions of real-world or imagined scenarios, each annotated with a corresponding misery score on a continuous scale from 0 (no misery) to 100 (extreme misery). These misery ratings represent subjective estimates of emotional distress associated with each event and were originally sourced from publicly available Misery Index blogs and user-curated compilations. Notably, the data was aggregated from three primary sources: the Misery Index blog curated by Bobby MGSK [14], a consolidated dataset available on Jericho Blog [13], and an associated Google Spreadsheet containing structured entries used as the basis for this study. We use a 0 to 100 scale to allow finer differentiation between different levels of emotional distress. A coarse scale like "low/medium/high" may not capture the small but meaningful differences in how miserable different situations feel. The continuous scale makes it possible to measure distress more precisely and is also useful in regression tasks. This design choice for further analysis follows the original data sources, which used similar scales for human ratings.

Each record consists of a short English-language description of a scenario, such as "Breaking a bone" or "Getting fired from a job," and a numeric label indicating its misery level. The text entries remain semantically diverse, encompassing a wide vari-ety of emotional contexts, including physical injury, social embarrassment, legal trouble, and medical emergencies. To retain the original intent and emotional texture of each description, preprocessing was kept minimal. Only superficial formatting corrections such as whitespace trimming and numerical conversions were applied, while stopword removal or semantic normalization was deliberately avoided.

The misery ratings are approximately symmetrically distributed, with a mean of 56.45 and a standard deviation of 17.59. Scores range from a minimum of 11 to a maximum of 100, with the 25th, 50th, and 75th percentiles falling at 43, 56, and 69, respectively. This suggests that most statements induce moderate to high levels of perceived misery, though both low-severity and extreme-severity cases are well represented.

To better understand the coverage and diversity of the dataset, each statement was manually assigned to one of ten high-level event categories based on its main theme. This categorization was used only for initial inspection of the dataset and was not involved in any of the experiments or analysis presented in the paper. These categories include Family or Relationship Issues, Accidents or Mishaps, Animal-related Incidents, Medical Emergencies, Embarrassment, Physical Injury, Crime or Legal Trouble, Professional or Work-related Problems, Gross/Disgusting Events, and an Other/Miscellaneous category. The most common category was Other/Miscellaneous, comprising 26.4% of all examples, followed by Family/Relationship Issues (16.3%) and Accidents/Mishaps (15.3%). Less frequent classes, such as Crime, Workplace issues, and Gross events, contributed fewer than 5% each. The long-tailed distribution of event types reflects the wide range of emotionally salient life situations considered in this dataset. This manually curated categorization enables more structured evaluation of model performance across semantic subgroups and provides a foundation for analyzing which types of misery are most challenging for LLMs to predict. It also informs downstream experiments involving BERT-based retrieval [11] and gamified reasoning with feedback-augmented LLMs, each of which relies on understanding event structure and affective content.

## 3   Conventional Benchmarking of Prompting Strategies

In this section, we evaluate various prompting strategies for predicting misery scores using large language models (LLMs). The goal is to benchmark conventional approaches under a unified regression framework, focusing on how different prompt types and sampling techniques influence prediction quality across several metrics.

### 3.1   Problem Formulation

This work addresses the task of predicting a numerical misery score from a natural language description of a life event. Formally, the objective is to learn a mapping from a text input x to a scalar output y$\in$ [0,100], where y reflects the perceived emo-tional distress caused by the event. The problem is framed as a supervised regression task, where the model estimates y'=f(x), and performance is assessed by comparing the predicted score y' to the ground truth y.

### 3.2 Language Model Architecture and Access

We utilize several commercially available Large Language Models (LLMs) accessed via API, including GPT-3.5, GPT-4, GPT-4o, and Azure ChatGPT [8,9,10]. Azure ChatGPT is a Microsoft-hosted deployment of GPT-4 provided through the Azure OpenAI Service. It offers the same underlying model as OpenAI's GPT-4 but with enterprise-grade deployment, regional endpoints, and performance monitoring suited for scalable experimentation [18]. These models are treated as black-box predictors, with no internal weight modification or fine-tuning. All predictions are generated through prompt engineering, exploring multiple prompting strategies to elicit numeric predictions. The choice of models reflects a range of instruction-following and few-shot generalization capabilities observed in prior evaluations [2,10].

### 3.3 Prompting Strategies

We explore three core prompting paradigms—zero-shot, few-shot, and reasoning-enhanced prompting—each aiming to evaluate a different aspect of LLM capability.

In the zero-shot setting, the model is provided only with a natural language description of the event and a simple instruction prompt to return a misery score. This approach evaluates the model's intrinsic generalization capability without exposure to any labeled examples, as previously explored in LLM survey benchmarks [12].

To encourage structured reasoning, we also employ a two-pass Chain-of-Thought (CoT) prompting strategy. In the first pass, the model is instructed to generate an intermediate reasoning process, describing why the event may be distressing. In the second pass, this reasoning output is supplied back to the model with a follow-up prompt to produce a final misery score. This staged approach aims to improve interpretability and decision alignment, though at the cost of increased latency [16].

In the few-shot setting, the model is given a small number of labeled examples (statement–score pairs) prior to the test instance. We compare three variations: (i) fixed prompting, where a static set of k examples is reused across predictions; (ii) random prompting, where a different set of k examples is sampled per instance; and (iii) embedding-based retrieval, where BERT-based sentence embeddings are used to retrieve semantically similar examples to inform the prompt dynamically [11]. This retrieval mechanism enables contextual relevance in the prompt, improving alignment for semantically similar inputs.

### 3.4 Evaluation Metrics

Model performance is quantitatively evaluated using Mean Absolute Error (MAE), Root Mean Squared Error (RMSE), Pearson correlation, Spearman rank correlation, and the coefficient of determination (R-squared). These metrics jointly capture prediction accuracy, linear alignment, ordinal consistency, and explained variance. MAE serves as the primary evaluation metric due to its direct interpretability in real-world contexts. Similar evaluation schemes have been applied in affective modeling and LLM regression tasks [4,5].

### 3.5 Prompting Strategy Evaluation

**Table 1.** Performance of different prompting strategies (zero-shot, CoT, few-shot with fixed or embedding-based examples) on the simple misery regression task (standard non-game-show setting). Subset of results shown for selected values of k (1, 2, 5) to highlight key performance trends.

| Metric | Zero-Shot Prompting | 2-Stage CoT Prompting | Fixed Samples | | | Embedding-based | | |
|---|---|---|---|---|---|---|---|---|
| | | | k=1 | k=2 | k=5 | k=1 | k=2 | k=5 |
| Mean Absolute Error (MAE) | 23.4771 | 24.2021 | 12.9875 | 12.485 | 14.42 | 12.422 | 13.408 | **12.3** |
| Root Mean Squared Error (RMSE) | 27.285 | 28.3047 | 16.847 | 16.503 | 17.95 | 16.227 | 16.904 | **15.97** |
| Pearson Correlation | 0.4511 | 0.4488 | 0.586 | **0.605** | 0.521 | 0.538 | 0.51 | 0.534 |
| Spearman's Rank Correlation | 0.5162 | 0.4909 | 0.588 | **0.617** | 0.573 | 0.551 | 0.514 | 0.532 |
| R-squared ($R^2$) | 0.201 | 0.2014 | 0.081 | 0.118 | -0.043 | 0.147 | 0.075 | **0.175** |

We begin by evaluating the scalar regression capabilities of large language models (LLMs) across different prompting strategies. As discussed in section [3.1], the model is provided with a textual description of an unusual or distressing real-life event and asked to predict a misery score—a scalar value ranging from 0 (least miserable) to 100 (most miserable). Each event is paired with a ground-truth misery score allowing us to quantitatively evaluate model predictions using standard regression metrics. We present the performance across multiple metrics, including MAE, RMSE, Pearson correlation, Spearman correlation, and R-squared, enabling a multifaceted view of prediction behavior. Results reveal that embedding-based few-shot prompting generally yields superior performance across metrics, consistent with prior findings on the value of semantically coherent prompting [3,11].

The zero-shot prompting baseline yields an MAE of 23.48 and RMSE of 27.29, establishing a meaningful benchmark without in-context examples. Correlation scores (Pearson: 0.4511, Spearman: 0.5162) suggest moderate alignment with human ratings, consistent with prior evaluations of GPT-3.5 and GPT-4 [10,12].

Adding reasoning via two-stage Chain-of-Thought (CoT) prompting does not improve results meaningfully (MAE: 24.20, RMSE: 28.30, R-squared: 0.2014). This aligns with findings that CoT helps in structured tasks but offers limited benefit in subjective settings [4,16].

Few-shot prompting offers clear gains. With just one example (k=1), MAE drops to 12.99, and further to 12.49 at k=2, with the highest Pearson correlation (0.606). However, gains plateau at higher k; at k=5, R-squared turns negative (–0.043), suggesting reduced diversity or overfitting [2,3,10].

Embedding-based prompting using BERT similarity [11] performs comparably (MAE: 12.30, RMSE: 15.97 at k=5), with R-squared: 0.175—higher than fixed-k—highlighting the advantage of semantically aligned examples [4,11].

Overall, few-shot and embedding-guided strategies outperform zero-shot and CoT, confirming the value of **contextual prompting** in our task.

## 4 Gamified Evaluation: Misery Game Show Simulation

To complement standard regression-based evaluation and explore the adaptive reasoning capabilities of Large Language Models (LLMs), we introduce a gamified testing environment termed the *Misery Game Show Simulation*. This evaluation framework draws inspiration from the format of the television game show *The Misery Index*, where human participants are tasked with judging the relative severity of bizarre or distressing real-life events [7]. Our simulation operationalizes this concept in a structured and controlled setting, enabling a more nuanced and interpretable assessment of LLM performance.

The framework incorporates a series of tasks that span comparative, ordinal, and scalar reasoning formats, each designed to test the model's ability to understand, compare, and quantify emotional intensity in natural language descriptions of life events. This approach is particularly well-suited to exploring affective cognition and context sensitivity in generative models, as emphasized in recent literature on fine-grained emotional reasoning in LLMs [3,4].

### 4.1 Simulation Design

The simulation is structured into a sequence of episodes, each consisting of multiple rounds that represent different cognitive subtasks. Across these rounds, the model is queried using structured prompts and evaluated on its ability to make ordinal, binary, and numeric predictions. Ground-truth misery scores are provided from the dataset described in Section 3 [13,14]. Two gameplay modes are implemented: one in which the model receives no corrective information across rounds (static mode), and one in which feedback is incorporated after each prediction (adaptive mode). The simulation thereby allows for the investigation of feedback-induced learning dynamics in black-box LLMs [5,17].

### 4.2 Game Rounds

Each episode in the Misery Game Show Simulation consists of a structured sequence of four rounds, with each round designed to test a specific aspect of emotional reasoning. Importantly, several rounds include multiple sub-questions, and selected questions incorporate feedback from earlier responses to simulate adaptive learning.

1. **Round 1**: Misery Lane (Ordinal Reasoning with Feedback)
   This round evaluates ordinal reasoning using two known reference events (1_base_1, 1_base_2) with fixed misery scores. The model answers two questions per episode:
   – **Q1 (1_1)**: The model is shown the two reference events and a target event (1_1). It must classify the target as {{{above}}}, {{{below}}}, or {{{between}}} the reference anchors. Correctness is evaluated based on how the actual score of 1_1 compares to those of the two anchors.
   – **Q2 (1_2)**: Before answering, the model receives feedback on whether its previous response was correct. Then, given a new target event (1_2) and the same two anchors, it must again classify it as above/below/between. This second question evaluates the model's ability to incorporate feedback to refine reasoning.
2. **Round 2**: More or Less Miserable (Binary Comparison with Feedback)
   This round probes binary emotional comparisons and introduces two questions per episode:
   – **Q3 (2_1)**: The model is given a reference story (2_1_base) with a known misery score and a second story (2_1) with a hidden score. It must decide whether the target is {{{higher}}} or {{{lower}}} in misery.
   – **Q4 (2_2)**: Feedback on Q3 is provided, indicating whether the comparison was accurate. The model is then presented with a new pair (2_2_base, 2_2) and repeats the binary comparison task. This assesses its ability to adjust judgment based on previous feedback.
3. **Round 3:** Master of Misery (Scalar Prediction)
   – **Q5:** The model is presented with a single unseen story and is required to output a scalar misery score between 1 and 100. This question forms the scalar regression baseline, and performance is assessed using the absolute error between the predicted and true scores.

4. **Bonus Round:** Margin of Misery (Interval Calibration)

This round tests the model's ability to localize uncertainty and produce bounded interval estimates around the correct score:

— **Q6 (4_1):** Predict a 30-point interval (±15 range), e.g., [35, 65], which must contain the true score.
— **Q7 (4_2)**: Predict a 20-point interval (±10 range).
— **Q8 (4_3)**: Predict a narrow 10-point interval (±5 range).

Each interval must exactly match the required width and contain the ground-truth score to be marked correct. Increasing difficulty across these sub-tasks evaluates calibration under tightening constraints.

The different rounds in the Misery Game Show are designed to test different types of emotional reasoning:

- **Round 1 (Ordinal reasoning)** checks if the model can place an event between two known levels of misery. This simulates how people compare events without knowing exact scores.
- **Round 2 (Binary comparison)** tests simpler pairwise judgments (e.g., "Is this more miserable than that?"), which are useful in everyday decisions.
- **Round 3 (Scalar prediction)** directly checks how accurately the model can assign a misery score.
- **Bonus Round (Interval prediction)** tests whether the model can estimate a score with uncertainty, which is important for cautious decision-making.

Each round captures a different aspect of emotional judgment. Together, they help evaluate how well LLMs can reason about emotional intensity in various formats.

All responses are automatically parsed and evaluated. Robust formatting constraints and answer validation logic are implemented to ensure consistency and minimize ambiguity across episodes.

## 4.3 Feedback and Adaptation

The adaptive gameplay mode introduces feedback after each round, wherein the correct answer is revealed to the model before proceeding to the next task. This mode is designed to test the model's ability to revise its internal reasoning and calibrate predictions based on recent correctness. The presence or absence of feedback serves as a key experimental variable for evaluating LLM adaptability under constrained memory and prompt length.

**Table 2.** Performance metrics of the model with and without feedback across rounds.

| Metric | Without_Feedback | With_Feedback |
|---|---|---|
| Round_1 | **54.41** | 38.16 |
| Round_2 | 72.06 | **77.63** |
| Bonus_Round | 45.10 | **50.88** |
| Overall | **55.46** | 54.89 |
| Avg_Distance_in_Round_3 | 23.41 | **17.82** |

Table 2 presents a comparative evaluation of the model's performance with and without feedback across all game rounds. As observed, the incorporation of feedback leads to notable improvements in Round 2 and the Bonus Round, suggesting enhanced contextual and comparative reasoning when iterative refinement is enabled. Furthermore, the average distance in Round 3—a proxy for proximity-based accuracy—shows a significant reduction (from 23.41 to 17.82) with feedback, indicating better calibration in numerical estimation tasks. Although a slight decline is observed in Round 1 performance with feedback, the overall performance remains comparable, with logical gains in tasks that benefit from cumulative context. These trends collectively support the hypothesis that feedback enables more adaptive and informed responses, aligning with the expected benefits of interactive learning.

**4.4 Misery Game Show Results (With Feedback)**

Models were invoked through their respective APIs (OpenAI, Azure, or Google VertexAI endpoints), with minor variations in system prompt formatting where required. The simulations were executed with random seeds 12, 123, and 1234 to test reproducibility and consistency across sampling variations.

A summary of progress is presented below, indicating successful completion of game simulations (✓) and failures or non-executions (×) per model and seed configuration.

**Table 3.** Model execution consistency across random seeds. GPT-family models exhibited stable performance across all seeds, whereas experimental models such as o1 and Gemini failed to execute under current pipeline configurations.

| Seed | GPT-3.5-turbo | GPT-4 | GPT-4-turbo | GPT-4o-mini | GPT-4o | o1-preview | o1-mini | Gemini-1.5-pro | Azure Chat |
|---|---|---|---|---|---|---|---|---|---|
| Seed 12 | ✓ | ✓ | ✓ | ✓ | ✓ | × | × | × | ✓ |
| Seed 123 | ✓ | × | ✓ | ✓ | ✓ | × | × | × | ✓ |
| Seed 1234 | ✓ | ✓ | ✓ | ✓ | ✓ | × | × | × | ✓ |

The Misery Game Show simulation evaluates the reasoning performance of language models in a structured, multi-round setting designed to mimic ordinal, binary, and scalar decision-making under uncertainty. The average accuracies across rounds and models are summarized in Table 4.

Table 4. Model-wise accuracy and prediction error across evaluation rounds.

| Model | Round 1 Accuracy (%) | Round 2 Accuracy (%) | Bonus Round Accuracy (%) | Overall Accuracy (%) | Avg. Distance in Round 3 |
|---|---|---|---|---|---|
| gpt-3.5-turbo | 42.92 | 74.58 | 28.33 | 45.71 | 30.00 |
| gpt-4-turbo | 50.00 | 70.83 | 50.56 | 56.19 | **15.13** |
| gpt-4o-mini | 55.83 | 70.00 | 28.06 | 47.98 | 20.90 |
| gpt-4o | **57.08** | 76.25 | **55.28** | **61.79** | 16.90 |
| azure chat | 38.16 | **77.63** | 50.88 | 54.89 | 17.82 |

Round-Level Performance Insights
Among the three game rounds, binary comparisons in Round 2 were the easiest for language models, with an average accuracy of 74.9%. This suggests that LLMs are particularly effective at making relative judgments when tasks are framed as direct comparisons. In contrast, Round 1, which involved three-way ordinal classification ("below," "between," "above"), and the Bonus Round, which required precise range estimation, were more challenging—achieving only 48.6% and 43.1% average accuracy, respectively. These lower scores reflect the greater difficulty LLMs face in categorical boundary reasoning and uncertainty estimation.

Model-Level Performance Summary
GPT-4o outperformed all other models with the highest aggregate accuracy (61.79%) and the lowest mean error (16.90) in direct score prediction. GPT-4-turbo and Azure ChatGPT followed with 56.19% and 54.89% accuracy, respectively. Notably, Azure ChatGPT slightly surpassed GPT-4o in binary comparisons (Round 2), indicating model-specific strengths depending on task structure.

Overall Implications
These results underscore the value of the Misery Game Show framework in revealing nuanced differences between models that might be missed under scalar-only evaluations. The performance drop in reasoning-intensive tasks highlights key areas where LLMs could benefit from targeted improvements, such as calibration or memory-based adaptation.

## 5    Limitations and Future Work

While this work explores the task of misery score prediction using large language models, there are some limitations that can be addressed in future work.
First, the dataset used in this study contains only 516 examples, which may limit the generalizability of the findings. Expanding the dataset and performing detailed error analysis could provide deeper insights into model behavior and failure cases. Second, the experiments were conducted exclusively with GPT-based models. Broader model comparisons—including other commercial or open-source LLMs—would allow for a more comprehensive understanding of model capabilities in affective prediction tasks. Third, while the results are consistent across models and prompting strategies, we did not apply statistical significance tests to the reported metrics. Including such tests, such as paired t-tests or bootstrapping, in future work could strengthen the analysis and provide additional confidence in the observed trends. Finally, predicting emotional distress from natural language involves ethical considerations. Incorrect predictions, especially underestimation of severe cases can have negative consequences in sensitive contexts such as mental health monitoring. This system is intended purely for research purposes, and any real-world deployment should include human oversight and be guided by appropriate ethical safeguards.

## 6    Conclusion

This study demonstrates the viability of using Large Language Models (LLMs) for predicting human-perceived misery scores from textual descriptions of life events. Through extensive experimentation with prompting strategies, we find that few-shot and retrieval-augmented prompting significantly enhance performance over zero-shot and reasoning-only baselines. The ability of LLMs to infer emotional severity from sparse supervision underscores their generalization capabilities in affective regression tasks.

Among the models evaluated, GPT-4o and GPT-4-turbo consistently yield the highest performance across scalar regression and structured game-based evaluations. These models achieve strong correlation with ground truth scores and demonstrate robust reasoning under feedback-driven settings. In contrast, smaller models such as GPT-3.5-turbo exhibit higher variance and poorer calibration, particularly under multi-way classification and numeric estimation tasks.

The structured *Misery Game Show* simulation provides additional insight into model capabilities across reasoning modalities. Binary comparisons emerge as the most tractable task for all models, with accuracies exceeding 70%, while ordinal and scalar prediction tasks remain more error-prone. Feedback incorporation consistently improves downstream performance, suggesting that LLMs are capable of adapting predictions when contextual grounding or corrective signals are introduced.

Despite these advances, limitations persist. Fine-grained scalar predictions remain challenging, particularly under low-context or ambiguous inputs. Models often struggle with emotional nuance and tend to under- or overestimate the severity of border-line scenarios. These findings highlight the need for further adaptation, tuning, and interpretability efforts, especially when deploying such systems in high-stakes, human-centered applications.

Overall, the results suggest that LLMs possess substantial potential in modeling subjective emotional reasoning, though achieving clinically reliable predictions requires continued refinement in both prompting and model alignment strategies.